# Online Signature Verification based on the Lagrange formulation with 2D and 3D robotic models

Moises Diaz, Miguel A. Ferrer, Juan M. Gil, Rafael Rodriguez, Peirong Zhang, Lianwen Jin

## Abstract

Online Signature Verification commonly relies on function-based features, such as time-sampled horizontal and vertical coordinates, as well as the pressure exerted by the writer, obtained through a digitizer. Although inferring additional information about the writer's arm pose, kinematics, and dynamics based on digitizer data can be useful, it constitutes a challenge. In this paper, we tackle this challenge by proposing a new set of features based on the dynamics of online signatures. These new features are inferred through a Lagrangian formulation, obtaining the sequences of generalized coordinates and torques for 2D and 3D robotic arm models. By combining kinematic and dynamic robotic features, our results demonstrate their significant effectiveness for online automatic signature verification and achieving state-of-the-art results when integrated into deep learning models.

## 1. Introduction

To verify an individual's authorship automatically, a set of enrolled digitized signatures are matched with a questioned specimen. Based on this comparison, an automatic signature verifier (ASV) provides an output measure (or score), which is used to accept or reject the individual's claimed identity [6, 12]. The challenge of developing high- performing ASVs has attracted interest in both academia and industry, and extends to devising ever more efficient verifiers [6].

To measure robustness in ASV, the most common accuracy metric is the Equal Error Rate (EER), which calculates the rate at which the False Acceptance Rate (FAR) and False Rejection Rate (FRR) are equal. Note that a lower EER indicates a more robust system. Typically, skilled and random forgeries are the two types of impostors used to evaluate ASVs [34]. Furthermore, to assess and improve robustness, adversarial attacks and defenses are studied extensively in the field. In [25], a sophisticated adversarial attack is proposed, generating nearly imperceptible perturbations on sig- nature strokes, which significantly threaten the reliability of ASVs in distinguishing between genuine and forged signatures. Training the ASVs to detect these perturbations has been proposed as a countermeasure. Also, cross-domain testing strategies are suggested, where the ASVs are trained on one database and tested on another with different characteristics.

Furthermore, one strategy commonly followed in the literature is to propose novel features that can better recognize the differences between a genuine and a forged signature. Many of the approaches focus on data produced by the movement of the writing tool, specifically, the kinematics associated with the pen tip. For example, a commonly used approach involves calculating velocity based on either its module or its horizontal and vertical components, which are derived from the input signature trajectory.

### 1.1. Related works

Such features have been employed in signature verification over the years, as is evidenced by various other works (e.g. [21, 20]).

Due to the capacity of some digitizers, pressure $p(t)$ and pen tip angles (i.e. elevation and azimuth) have also been used in some works. However, the usefulness of these angles is not universally acknowledged. While some authors have found ways to use them effectively in verification, as can be seen in [28], for example, their quantized effect and noise often introduce confusion into the classifiers, as noted in [24, 7].

Some other authors have obtained promising results with other features that retain the physical meaning of pen-tip kinematics. One example of such features is acceleration, which can be observed either as a module of acceleration or as centripetal acceleration [23]. Other popular features include the geometrical tangential angle of the trajectory curve [23] and the curvature radius, which feature a physical significance related to the handwriting phenomenon. Table 1 includes examples of these types of function-based features.

Table 1: Common function-based features used in ASV, with physical significance and processed through $(x(t), y(t), p(t))$.

| | |
|---|---|
| Trajectory coordinates and the pen pressure: | $(x(t), y(t), p(t))$ |
| First-order derivatives: | $(\dot{x}(t), \dot{y}(t), \dot{p}(t))$ |
| Second-order derivatives: | $(\ddot{x}(t), \ddot{y}(t))$ |
| Module of velocity: | $v(t) = \sqrt{\dot{x}(t)^2 + \dot{y}(t)^2}$ |
| Path-tangent angle: | $\theta(t) = \arctan(\dot{y}(t)/\dot{x}(t))$ |
| $\cos(\theta(t)), \sin(\theta(t))$ | |
| First-order derivatives of $v(t)$ and $\theta(t)$: | $\dot{v}(t), \dot{\theta}(t)$ |
| Log curvature radius: | $\rho(t) = \log(v(t)/\theta(t))$ |
| Centripetal acceleration: | $c(t) = v(t) \cdot \theta(t)$ |
| Module of acceleration: | $a(t) = \sqrt{\dot{v}(t)^2 + c(t)^2}$ |
| Ratio of the minimum and the maximum speed over a window of 5 samples: | $v_i^5 = \min(v_{i-4}, \ldots, v_i)/\max(v_{i-4}, \ldots, v_i)$ |
| Stroke length to width ratio over a window of $n$ samples: | |
| $r_i^n = a_i/b_i$ where $a_i = \sum_{k=i-n}^{i} \sqrt{(x_k - x_{k-1})^2 + (y_k - y_{k-1})^2}$ | |
| and $b_i = \max(x_{i-n}, \ldots, x_n) - \min(x_{i-n}, \ldots, x_n)$ | |

In addition to some of these features, [26] includes the derivative of the path-tangent angle, the ratio of the minimum and maximum speeds over a window of 5 samples, the angle of $(x˙(t), y˙(t))$, and the cosine of the angle of consecutive samples. All these features entail a physical component and significance.

Another approach for estimating the curvature of the signature is presented in [18]. This method uses the geometric relationship between the salient points of the signature. Additionally, the authors formulate torsion as a feature that takes into account the trajectory of the extreme points (minima and maxima) and their neighbouring points.

An excellent example of work that uses most of these function-based features can be seen in the most recent signature verification competition held at ICDAR 2021 [34]. Relatedly, the use of robots in handwriting applications has been the subject of multifaceted studies. For example, robots have been used to assist children in improving their handwriting [19], while others have been trained to produce human-like handwriting [39]. The aesthetics and dynamics of handwriting produced by robots have also been studied [38, 31], considering forensic handwriting examiners' perspectives [11]. Recent research has also demonstrated that robots can be trained to produce deceptive signatures [2], which motivates further exploration of the use of robotic function-based features for signature verification. Two papers [7, 8] have focused on leveraging the kinematics of the IAB IRB120 industrial robot during signature execution. To this end, the angles in the robots' joints were calculated, taking into account their movement according to the motion of the pen tip. In incorporating such features, the authors found their performance to be competitive according to a benchmark analysis. However, other aspects related to robot movement [10], such as robot dynamics, were not covered.

Related existing works have shown promising results when investigating robot dynamics to model human movement. An example can be seen when modelling gait movements by applying a Lagrange formulation in [27].

The authors validated that the torque readings estimated in the ankle, knee, and hip were similar to those gauged by specific sensors. A proposal to obtain the torque numbers in the lower limb exoskeletons was studied in [1]. They demonstrated the possibility of tracking hip and knee torques with their model. With regard to the upper extremities, a solution involving torque control was proposed in [5], using a human musculoskeletal arm model. This was a 2 degrees-of-freedom model in which the author developed the Euler-Lagrangian formulation to obtain the torque figures. These works lead us to believe that the dynamics involved in handwriting production can be estimated using an associated robot and a Lagrange formulation to obtain the torques and the generalized coordinates. In a recent study [9], kinematic and dynamic robotic features were employed in automatic signature verification (ASV). Instead of approaches using Lagrangian formulas, this study used estimated kinematic and dynamic features derived from a multilayer perceptron neural network. The network was trained with kinematic and dynamic data measured from an industrial robot and achieved promising results.

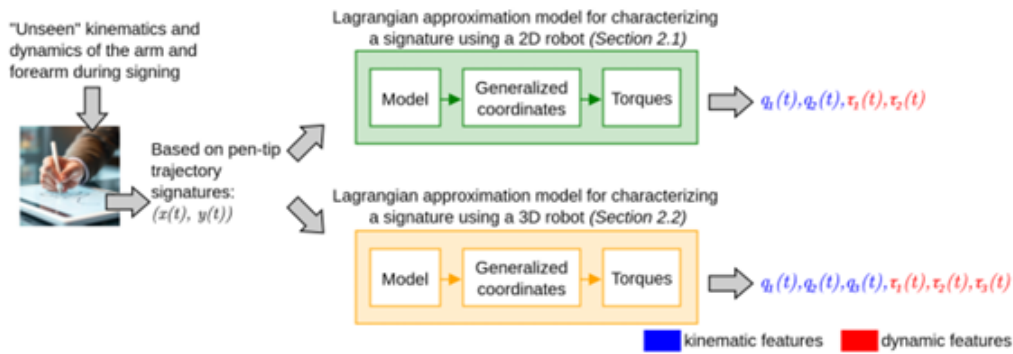

(a) Estimation of robotic features from the trajectory of an online signature.

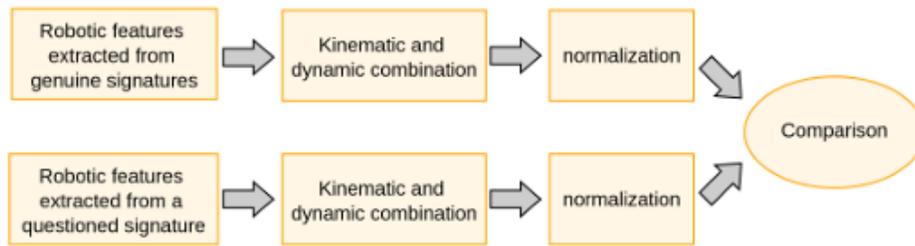

(b) Automatic signature verification using robotic features.

Figure 1: Diagram illustrating the overall process of robotic feature extraction from an online signature and its application in automatic signature verification.

*1.2. Contributions*

In this work, we propose a unique feature set for signature verification, derived from generalized coordinates and torques. This approach aims to replicate the signing process, providing the features with a physical factor and significance, inspired by the kinematics and dynamics of the human arm and forearm during signing. The method was tested on two robotic models, 2D and 3D. To achieve this, a set of non-linear trans formations based on the Lagrange formulation were performed. This mathematical procedure may also be useful for further advanced systems based on physical models. An overview illustrating the overall process of extracting robotic features and their use in ASV is provided in Figure 1.

Robotic features, i.e. generalized coordinates and torques, were combined in an extensive evaluation using multiple publicly available signature databases. These databases comprise signatures written in various scripts (Western, Bengali, Devanagari, etc.), and captured using different digitizing devices (Wacom tablet, ePad-ink tablet, smartphone, etc.).

After a study of the contribution of the new features over traditional machine learning classifiers and databases– given the fact that the state of the art in online sig nature verification has been consistently based on deep learning approaches in recent years– we incorporated our features into a state-of-the-art deep learning classifier [20] and carried out experiments using a common benchmark based on the DeepSignDB database, participating in the SVC-onGoing competition and achieving top-3 results. In summary, this study can provide new insights into the signing process by demonstrating the application of a physically meaningful set of features.

The rest of the paper is organized as follows: Section 2 presents the Lagrange formulation and the 2D and 3D robotic models used to deduce the generalized coordinates and torques during signing. Next, Section 3 describes the evaluation carried out to analyse the use of novel function-based features in online automatic signature verification. The performance analysis is provided in Section 4, using multiple signature databases with traditional machine-learning classifiers. In Section 5, we demonstrate that our features yielded top-3 state-of-the-art results when integrated into a deep learning sig nature verifier. Finally, the paper concludes with Section 6.

## 2. Lagrangian approximation model to characterize a signature

The Euler-Lagrange method is a systematic formulation based on energetic considerations, which solves the inverse dynamics challenge [4]. This approach determines the torques required to achieve the generalized coordinates, which can be computed using inverse kinematics [7] based on pen-tip trajectory signatures, $(x(t), y(t))$. To this end, we propose a 2D and a 3D robotic model, as shown in Figure 2. The 2D model is a rigid robot consisting of two links and two degrees of freedom, whereas the 3D robot has three links and three degrees of freedom.

*2.1. A 2D robotic model*

The inverse dynamics of a 2D rigid robot with n degrees of freedom are governed by the Lagrange formulation as follows [4]

$$\frac{d}{dt}\left(\frac{\partial L\left(q(t),\dot{q}(t)\right)}{\partial \dot{q}}\right) - \frac{\partial L\left(q(t),\dot{q}(t)\right)}{\partial q} = \tau, \tag{1}$$

where $q = [q_1(t), q_2(t), ..., q_n(t)]^T$ and $\dot{q} = [\dot{q}_1(t), \dot{q}_2(t), ..., \dot{q}_n(t)]^T$, respectively, denote the generalized coordinates and velocities of the robot links and $\tau = [\tau_1(t), \tau_2(t), ..., \tau_n(t)]^T$ is the vector of torques applied to the robot. $L(q(t), \dot{q}(t),)$ denotes the Lagrangian of the system, which is defined as the difference between the kinetic energy, $K(q(t),\dot{q})$, and the potential energy, $U(q(t))$, of the rigid robot:

$$L\left(q(t),\dot{q}(t)\right) = k\big(q(t),\dot{q}(t)\big) - U(q(t)) \tag{2}$$

To simplify the notation, we will not include the time dependence of generalized coordinates and velocities below. In order to obtain the kinetic energy of links 1 and 2 of this 2D robot, the velocities of the centres of mass of each link are considered:

$$\text{K}(v_C, q) = \left[\frac{1}{2}m_1 v_{c1}{}^2 + \frac{1}{2}I_1 \dot{q}_1{}^2\right] + \left[\frac{1}{2}m_2 v_{c2}{}^2 + \frac{1}{2}I_2(\dot{q}_1 + \dot{q}_2)^2\right] \tag{3}$$

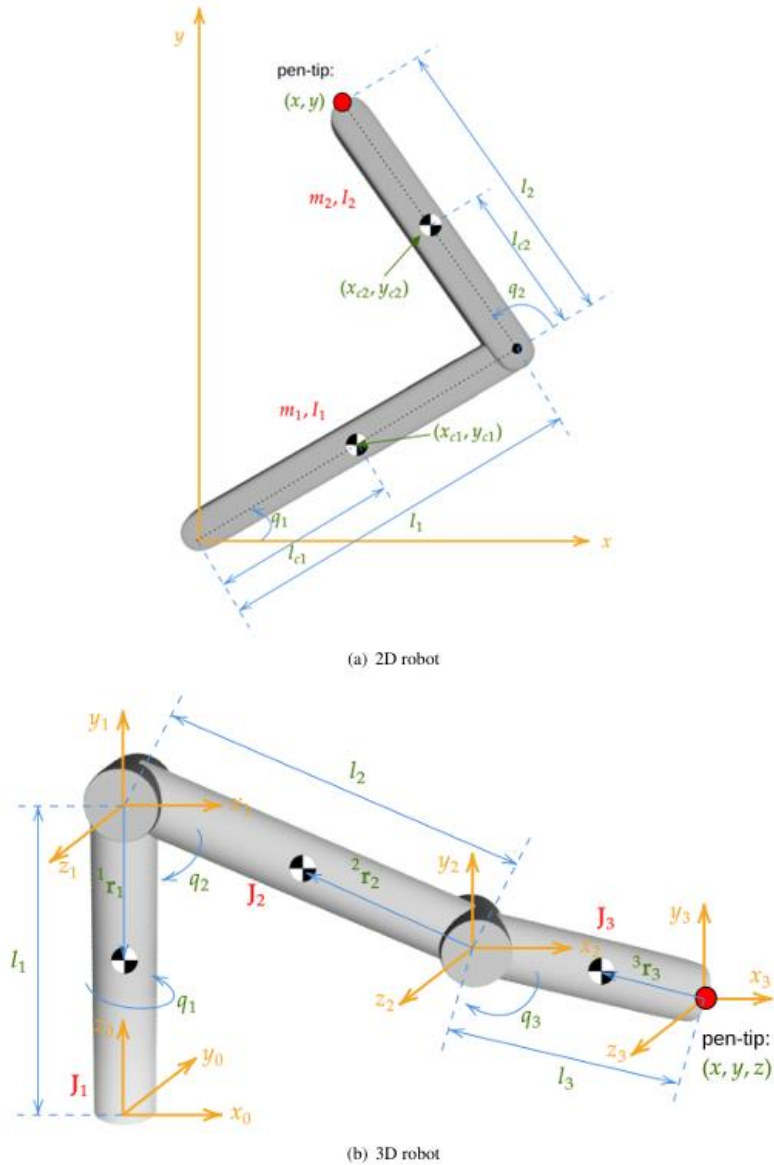

(a) 2D robot

(b) 3D robot

*Figure 2: Notation used for the 2D and 3D robotic models.*

where m1 and m2 denote the mass of the links, $I_1$ and $I_2$ are the moments of inertia of links 1 and 2, respectively, and $v_c = (v_{c1}\ v_{c2})$ denotes the velocities of the center of mass of the links, which are given in terms of Cartesian coordinates as $v_{c1} = [\dot{x}_{c1}\ \dot{y}_{c1}]^\mathrm{T}$ and $v_{c2} = [\dot{x}_{c2}\ \dot{y}_{c2}]^\mathrm{T}$. Taking into account the Cartesian coordinates of the centre of mass of links 1 and 2 in the X-Y plane, we obtain the following kinetic energies:

$$k_1\ (\dot{q}_1) = \tfrac{1}{2}\, m_1 I_{c1}{}^2\ \dot{q}_1{}^2 + \tfrac{1}{2} I_1\ \dot{q}_1{}^2$$
$$k_2\ (q_2, \dot{q}_1, \dot{q}_2) = \tfrac{1}{2}\, m_2 I_1{}^2\ \dot{q}_1{}^2 + \tfrac{1}{2}\, m_2 I_{c2}{}^2\ [\dot{q}_1{}^2 + 2\dot{q}_1\dot{q}_2 + \dot{q}_2{}^2] +$$
$$+\ m_2 I_1{}^2 I_{c2}{}^2 [\dot{q}_1{}^2 + \dot{q}_1\dot{q}_2]\ cos q_2 + \tfrac{1}{2} I_2 [\dot{q}_1{}^2 + \dot{q}_2{}^2] \tag{4}$$

The potential energy of each rigid robot link is:

$$U_1\ (q_1) = m_1 I_{c1} g cos(q_1)$$
$$U_2\ (q_1, q_2) = m_2 I_1 g cos(q_1) + m_2 I_{c2} g cos(q_1 + q_2) \tag{5}$$

The Lagrangian of the 2D robot is then expressed as:

$$L\ (q, \dot{q}\ ) = k_1\ (\dot{q}_1) + k_2\ (q_2, \dot{q}_1, \dot{q}_2) - U_1\ (q_1) - U_2\ (q_1, q_2) \tag{6}$$

Since the rigid robot has two degrees of freedom, the resulting equations of motion are (from Eq. (1)):

$$\frac{d}{dt}\Big( \frac{\partial L\ (q,\dot{q})}{\partial \dot{q}_1}\ \Big) - \frac{\partial L\ (q,\dot{q})}{\partial q_1} = \tau_1$$

$$\frac{d}{dt}\Big( \frac{\partial L\ (q,\dot{q})}{\partial \dot{q}_2}\ \Big) - \frac{\partial L\ (q,\dot{q})}{\partial q_2} = \tau_2 \tag{7}$$

Operating on the Lagrangian, the equations of motion can be written as follows:

$$\boldsymbol{\tau} = \mathbf{D(q)}\ddot{\boldsymbol{q}} + \mathbf{H(q, \ddot{q})} + \mathbf{C(q)} \tag{8}$$

where τ are the two torques applied to each link, $\mathbf{D(q)}$ the $n \times n$ inertial matrix, $\mathbf{q}^{..}$ the second derivates of the two generalized coordinates, $\mathbf{H(q, \ddot{q})}$ is the $n \times 1$ Coriolis matrix and $\mathbf{C(q)}$ the $n \times 1$ gravity matrix.

The relationship between the generalized coordinates, $q_1(t)$, $q_2(t)$, and the position of the final end-effector, $x(t)$, $y(t)$, which is registered in the signature databases, can be obtained from the homogeneous transformation between the coordinate frame of the base $\{S_0\}$ and the coordinate frame of the last link $\{S_2\}$:

$$^0T_2 = {}^0T_1 \cdot {}^1T_2, \tag{9}$$

where $^0T_1$ is the homogeneous matrix in 2D between coordinate frames $\{S_0\}$ and $\{S_1\}$, and $^1T_2$ is the homogeneous matrix in 2D between coordinate frames $\{S_1\}$ and $\{S_2\}$. Therefore, $^0T_2$ is given by:

$$^0T_2 = \begin{pmatrix} \cos q_{12} & -sin q_{12} & I_1 cos(q_1) + I_2 cos(q_{12}) \\ \sin q_{12} & \cos q_{12} & I_1 sin(q_1) + I_2 sin(q_{12}) \\ 0 & 0 & 1 \end{pmatrix} \tag{10}$$

The entries at positions (2,3) and (3,3) in $^0T_2$ represent $x(t)$ and $y(t)$, respectively. Using the law of cosines, the following expression is then obtained:

$$\cos(q_2(t)) = \frac{((x(t))^2 + (y(t))^2 - l_1{}^2 - l_2{}^2)}{2 l_1 l_2}, \tag{11}$$

which allows us to obtain $q_2$ from the $(x(t), y(t))$ coordinates. However, for the calculation of this generalized coordinate, it is more expedient, from a computational point of view, to use the following expression:

$$q_2(t) = atan\left(\frac{\pm\sqrt{1-\cos(q_2(t))}}{\cos(q_2(t))}\right). \tag{12}$$

This last equality provides two possible solutions, depending on whether the positive or negative sign is used. The former corresponds to the "elbow up" configuration of the robot, while the latter corresponds to the "elbow down" configuration. For ergonomic reasons, we used the positive solution in this work. Finally, the generalized coordinate $q_1$ is obtained from the following equation:

Table 2: Denavit-Hartenberg parameters for the 3D robot (Figure 2-b).

| Joint $k$ | $\delta_k$ | $d_k$ | $a_k$ | $\alpha_k$ |
|-----------|------------|-------|-------|------------|
| 1 | $q_1$ | $l_1$ | 0 | $\pi/2$ |
| 2 | $q_2$ | 0 | $l_2$ | 0 |
| 3 | $q_3$ | 0 | $l_3$ | 0 |

$$q_1(t) = atan\left(\frac{y(t)}{x(t)}\right) - atan\left(\frac{l_2 sin(q_2(t))}{l_1 + l_2\, cos(q_2(t))}\right) \tag{13}$$

Given that the ergonomic position of the robot has been considered, the initial values of the generalized coordinates should be $q_1(0) = \pi/4$ rad and $q_2(0) = \pi/2$ rad. Using direct kinematics, the starting point of the signature (for all the databases analysed) can be determined. Finally, the trajectory of the signature is then shifted according to the starting point.

It is worth mentioning that the $(x(t), y(t))$ coordinates should be provided according to the International System of Units. To this end, they must be transformed using the resolution of the digitizer, measured in dots per inch (dpi). The specific database determines this resolution, as outlined in Table 4. Similarly, the temporal sequence should be expressed in seconds.

Conversely, the numerical values of the parameters for the 2D robot, denoted as $l_1, l_2, l_{c1}, l_{c2}, m_1, m_2, I_1, I_2$ and illustrated in Figure 2-a, represent the average values for humanoid arms. These values were derived from the study conducted in [3]. In particular, the first and second links have masses of $m_1 = 1.8425$ kg and $m_2 = 1.1132$ kg, respectively, while their moments of inertia are $I_1 = 0.0133$ kg m$^2$ and $I_2 = 0.0021$ kg m$^2$. The lengths of the limbs are $l_1 = 0.2820$ m and $l_2 = 0.2643$ m, respectively, and their centers of mass are located at $l_{c1} = 0.1447$ m and $l_{c2} = 0.1090$ m.

## 2.2. A 3D robotic model

In this section we formulate the inverse dynamics of a 3D robot using the Lagrange method. One approach to implementing the Lagrangian formulation involves utilizing homogeneous transformation matrices. In a previous study [37], an algorithm based on these matrices and the Denavit-Hartenberg (DH) parameters [36] was proposed.

---

**Algorithm 1** Lagrange computational algorithm for the dynamic model of the 3D robot.

**Input:** Generalized coordinates $\mathbf{q}$, DH parameters, and dynamic properties (link lengths ($l_i$), link masses ($m_i$), link COM positions ($^i\mathbf{r}_i$), link inertias ($I_{x,y,i}$)).

**Output:** Torques $\tau$

1: $^0T_1, {}^1T_2, {}^2T_3$       ▷ Homogeneous transformation matrices
2: $^0T_i$       ▷ Concatenated matrices
3: $\mathbf{a} = [0, 0, -9.81]$       ▷ Gravitational acceleration
4: $\mathbf{U}_{i,j} = \frac{\partial {}^0 T_i}{\partial q_j}$    $\mathbf{U}_{i,j,k} = \frac{\partial U_{i,j}}{\partial q_k}$       ▷ Inter-link interation effects
5: $\mathbf{J}_i = \text{pseudoinertial\_matrix}(I_{x,y,i}, m_i, {}^i\mathbf{r}_i)$       ▷ pseudoinertial matrix for each link
6: $\mathbf{D}(\mathbf{q}) = d_{i,j} = \sum_{k=(\max i,j)}^{n} \text{trace}\left(\mathbf{U}_{kj}\mathbf{J}_k\mathbf{U}_{ki}^T\right)$       ▷ Inertial matrix
7: $\mathbf{H}(\mathbf{q}, \dot{\mathbf{q}}) = h_i = \sum_{k=1}^{n} \sum_{m=1}^{n} h_{ikm}\dot{q}_k\dot{q}_m$       ▷ Coriolis and centripetal matrix
8:      where: $h_{ikm} = \sum_{j=\max(i,k,m)}^{n} \text{trace}\left(\mathbf{U}_{jkm}\mathbf{J}_j\mathbf{U}_{ji}^T\right)$
9: $\mathbf{C}(\mathbf{q}) = c_i = \sum_{j=1}^{n}(-m_j\mathbf{a}\mathbf{U}_{ji}{}^j\mathbf{r}_j)$       ▷ Gravity matrix
10: $\tau = \mathbf{D}(\mathbf{q})\ddot{\mathbf{q}} + \mathbf{H}(\mathbf{q}, \dot{\mathbf{q}}) + \mathbf{C}(\mathbf{q})$       ▷ Torque matrix

---

We have followed the algorithm proposed by Uicker [37] to obtain the dynamic model of the 3D robot through the Lagrange procedure, which is summarized in Algorithm 1. This approach involves using the matrices $^{i-1}T_i$, which establish relationships between consecutive coordinate frames of the 3D robot. To construct these matrices, the DH parameters, provided in Table 2, are derived from our 3D robotic model, illustrated in Figure 2-b.

First, a coordinate frame is assigned to each link, according to the DH parameters. Then, the rotational and positional relationships between two consecutive links are calculated using the homogeneous transformation matrices: $^0T_1, {}^1T_2, {}^2T_3$. These matrices facilitate the calculation of the concatenated matrices, $^0T_i$.

We then calculate the dual inter-link interaction effects. The effect of the movement of one joint on another joint is calculated by $\mathbf{U}_{i,j} = \frac{\partial {}^0 T_i}{\partial q_j}$, while the effect of the movement of two joints on another joint is calculated as: $\mathbf{U}_{i,j,k} = \frac{\partial U_{i,j}}{\partial q_j}$. Next, we compute the pseudoinertial matrices for each link, Ji. They include the moments of inertia, products of inertia, mass, and center of mass coordinates of each link. For the sake of simplicity, we have assumed that all the links are thin cylinders with zero cross inertias, as illustrated in Figure 2-b.

We then computed the three dynamic coefficients needed to calculate the torques in the joints. First, the elements of the inertial matrix, $\mathbf{D}(\mathbf{q})$, are defined by the trace of the following operation:

$$\mathbf{D}(\mathbf{q}) = d_{ij} = \sum_{k=(\max i,j)}^{n} trace(U_{kj}J_k U_{ki}^T) \tag{14}$$

Second, the effects of the Coriolis and centripetal forces, $\mathbf{H}(\mathbf{q},\dot{\mathbf{q}})$, are calculated as follows:

$$\mathbf{H}(\mathbf{q},\dot{\mathbf{q}}) = \mathrm{h}_i = \sum_{k=1}^{n} \sum_{m=1}^{n} h_{ikm}\,\dot{q}_k \dot{q}_m \qquad (15)$$

Where $\mathrm{h}_{ikm} = \sum_{j=\max(i,k,m)}^{n} trace(\boldsymbol{U}_{jkm}\boldsymbol{J}_j\,\boldsymbol{U}_{ji}^T)$. Finally, the gravity matrix, $\mathbf{C}(\mathbf{q})$, is expressed, whose elements are defined by:

$$\mathbf{C}(\mathbf{q}) = \mathrm{C}_i = \sum_{j=1}^{n} trace(-m_j a \boldsymbol{U}_{ji}\,{}^j \boldsymbol{r}_j) \qquad (16)$$

where $\mathbf{a}$ is the acceleration due to gravity in the $\{S_0\}$ coordinate frame, and ${}^j\mathbf{r}_j$ denotes the homogeneous coordinate vector of the centre of mass (COM) of link $j$ expressed in the coordinate frame of link $j$. Similar to the 2D robot, we calculated the dynamic equation of the system, already provided in Eq. (8).

To calculate the torques, we need the generalized coordinates. Following a procedure similar to that outlined in [7], we can directly compute the value of $q_1(t)$ based on the geometry of the 3D robot: $q_1(t)$ as:

$q_1(t) = \mathrm{atan}(y(t)/x(t))$, with reference to frame $\{S_0\}$. The values of $q_2(t)$ and $q_3(t)$ were determined as follows:

$$q_2(t) = \mathrm{atan}\left(\frac{z(t) - l_1}{\sqrt{x(t)^2 + y(t)^2}}\right) - atan\left(\frac{l_3 sin(q_3(t))}{l_2 + l_3\,cos(q_3(t))}\right)$$

$$q_3(t) = \mathrm{acos}\left(\frac{x(t)^2 + y(t)^2 + (z(t) - l_1 - l_2{}^2 - l_3{}^2)}{2 l_2 l_3}\right) \qquad (17)$$

It is worth pointing out that $(x(t), y(t))$ were obtained using the digitizer, while $z(t)$ remained fixed at a certain value. To reference these values to the $\{S_0\}$ coordinate frame, we consistently defined the starting point for each signature. To achieve this, we heuristically determined the offset as $(p_m + 0.35\,(p_M - p_m), 0, 0.3\,l_1)$ relative to the base coordinate frame, $\{S_0\}$, ensuring that the signature could be written within the robot's working area. Here, $p_m = \mathrm{l}_3\,\sin(\mathrm{acos}((\mathrm{l}_1 - \mathrm{offset}_z - \mathrm{l}_2/\mathrm{l}_3))$, and $p_M = \sqrt{(\mathrm{l}_2{}^2 + \mathrm{l}_3{}^2) - (\mathrm{l}_1 - \mathrm{offset}_z)^2}$. Applying direct kinematics, we computed the generalized coordinates for the initial starting point.

Similar numerical values used for the 2D robot were also applied to this robot. Specifically, the masses of the links were $m_1 = 33.9458$ kg, $m_2 = 1.8425$ kg, and $m_3 = 1.1132$ kg, while the lengths were $l_1 = 0.6644$ m $l_2 = 0.2820$ m, and $l_3 = 0.2643$ m.

Finally, Table 3 presents examples of torques calculated using the 2D robot and 3D robot for different types of signatures. In signatures that include text, an ascending pattern with a positive slope can be observed in the $x(t)$ coordinate due to left-to-right writing, alongside an oscillatory movement in the vertical coordinate, $y(t)$. However, this effect is not reflected in the torque sequences, which exhibit oscillatory patterns in all cases due to the angular accelerations in each link. A brief analysis of these torque features is shown in Figure 2, where it can be observed that torque $\tau_1$ in the 2D robot is consistently greater than torque $\tau_2$ for all analysed signatures, both genuine and forged. Similarly, in the case of the 3D robot, torque $\tau_3$ exhibits behavior analogous to $\tau_2$ in the 2D robot, highlighting the significance of torque $\tau_1$ associated with the newly added link. Consistent with the physics of movement, where torque is defined as force multiplied by distance, the baseline values follow the order $\tau_1 > \tau_2 > \tau_3$ for both robots. These torques fall within different ranges of *Nm*, determined by the mass being moved and the distance of application in each link. Another notable observation is the non-linear relationship between the trajectory and the movement dynamics, suggesting that torque measurements could provide valuable additional features for signature verification systems.

Table 3: Example of temporal torque sequences (in Nm) estimated from the 2D robot and the 3D robot for various signature styles. The temporal trajectory sequence is included to emphasize that the torque sequences were derived from these trajectories.

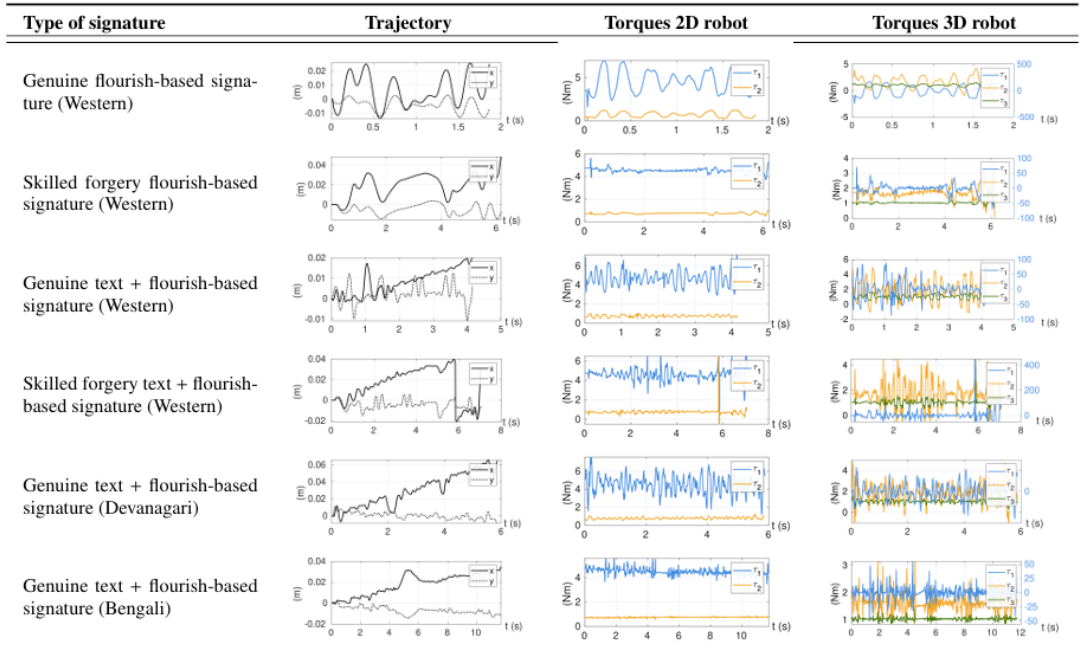

### 3. Evaluation methodology

Two evaluations were conducted. First, traditional machine-learning classifiers were employed to assess the effectiveness of generalized coordinates and torques in online ASV. Second, a deep-learning classifier was used to demonstrate the potential to achieve state-of-the-art performance with the features proposed.

*3.1. An evaluation using traditional machine-learning verifiers*

#### 3.1.1. *Databases*

Several signature databases were used to properly analyse newly proposed function- based features. The databases used include signatures in different scripts, such as Western, Bengali, and Devanagari, acquired using various devices in office and mobile phone scenarios, with different characteristics. A detailed description of these databases is presented in Table 4.

Almost all databases provide the spatiotemporal tuple ($x(t), y(t), p(t)$), while some also provide pen-ups and pen-downs without interruption during the transitions. It is important to note that only the $(x(t), y(t))$ coordinates were used to estimate the generalized coordinates and torques. When pen-ups were not included, skips produced by real pen-ups were omitted. The main motivation for using this set of databases was to quantify the automatic verification performance transparently in a wide range of cases.

Table 4: Summary of the databases evaluated.

| Database | Device type | Script | Users | Genuine | Forgeries | Resolution (dpi) | Rate (Hz) |
|---|---|---|---|---|---|---|---|
| | | *To the evaluation with traditional machine-learning ASVs* | | | | | |
| MCYT-100 [29][1] | Wacom Intuos A6 | Western | 100 | 2,500 | 2,500 | 2,540 | 100 |
| MCYT-330 [29][1] | Wacom Intuos A6 | Western | 330 | 8,250 | 8,250 | 2,540 | 100 |
| BiosecurID-SONOF [16][1] | Wacom Intuos3 A4 | Western | 132 | 2,112 | 1,584 | 2,540 | 100 |
| SUSIG-Visual [22][2] | ePad-ink tablet | Western | 94 | 1,880 | 940 | 13,087* | 70† |
| SUSIG-Blind [22][2] | Wacom's Graphire2 | Western | 88 | 820 | 880 | 1000 | 80† |
| mobile SG-NOTE [26][1] | Samsung Galaxy Note | Western | 25 | 500 | - | 2,540* | 70† |
| OnOffSigBengali-75 [13][3] | Wacom Intuos A3 | Bengali | 75 | 1800 | - | 2,540 | 100 |
| OnOffSigDevanagari-75 [13][3] | Wacom Intuos A3 | Devanagari | 75 | 1800 | - | 2,540 | 100 |
| | | *To the evaluation with deep-learning ASV* | | | | | |
| DeepSignDB [33][4] | 5 Wacom+3 Samsung | Western | 1,526 | 24,434 | 20,038 | NA | NA |
| SVC2021 EvalDB [35][5] | Wacom+94 smartphones | Western | 194 | 1,552 | 3,104 | NA | NA |

\* denotes resolution heuristically estimated. † denotes non-uniform sampling rate. NA: Not available due to device dependency.
The used databases can be downloaded from: [1]`http://atvs.ii.uam.es/atvs/databases.jsp`, [2]`https://biometrics.sabanciuniv.edu/susig.html`, [3]`https://gpds.ulpgc.es/`, [4]`atvs.ii.uam.es/atvs/DeepSignDB.html`, [5]`codalab.lisn.upsaclay.fr/competitions/9189`

### 3.1.2. Automatic signature classification

To analyse the versatility of using generalized coordinate and torque features, we employed two traditional machine-learning classifiers, each based on different methodologies. This approach allows for a comprehensive evaluation of signature properties, assessing robotic features in terms of their performance.
The first ASV is based on functions and Dynamic Time Warping (DTW) distance. Enrolled and questioned signatures are compared by using the same DTW configuration proposed in [15]. The feature vector was built using the generalized coordinate features or the torque-based features. Additionally, we also studied the performance when both generalized coordinates and torques are used as a feature vector. Next, the first- and second-order time derivatives are added to each feature vector and, finally, a z-score normalization is performed.

The second ASV is based on the histogram and a Manhattan distance, namely MAN-based ASV in this work. The feature vector consists of two histograms with absolute and relative frequencies, which were adapted to the features proposed. As results, a histogram for the torque-based features, $h_\tau$, and another for the generalized coordinates were developed. The similarity between the reference and questioned features is then obtained from the Manhattan distance [32].

### 3.1.3. Experimental protocol and metrics used

Three configurations of torques and generalized coordinates were tested to assess which is the best way of using them. In this way, we combined the generalized co-ordinates and torques with their derivates and between them as follows: Firstly, the feature sets $F_1$ and $F_2$ evaluated the generalized coordinates and torques individually. They can be defined as: $F_1 = (q_1, q_2, \dot{q}_1, \dot{q}_2, \ddot{q}_1, \ddot{q}_2)$ and $F_2 = (\tau_1, \tau_2, \dot{\tau}_1, \dot{\tau}_2, \ddot{\tau}_1, \ddot{\tau}_2))$. Secondly, we evaluated the combination of generalized coordinates and torques in $F_3$. For DTW-based ASV, $F_3 = [F_1, F_2]$, while for MAN-based ASV, $h = [h_q \parallel h_\tau]$.

We conducted experiments on skilled and random forgeries in biometrics by following a standard protocol [34, 7] and the common procedure in the domain [6]. In all the experiments, the first five reference signatures were chosen as the reference set. DET curves and the Equal Error Rate (EER) were then reported. All the databases were used for the random forgery scenario. For FRR, all genuine signatures, with the exception of those in the reference set, were used, while for FAR, one genuine signature of other users was used to generate the curve. For skilled forgeries, the FRR curve was the same, while the FAR curve was generated using all available skilled forgeries for each user.

### 3.2. An evaluation using deep-learning classification

### 3.2.1. Databases

The purpose of this evaluation is to compare our results with the state of the art. For a fair comparison, two initiatives have recently helped address this issue. First, the DeepSignDB [33] has emerged as a large database that enables meaningful statistical experiments in signature verification. Second, the SVC-onGoing, based on the ICDAR 2021 Competition on On-Line Signature Verification [35], is an open competition. The SVC2021 EvalDB database was developed for this competition and was used in this article. Table 4 includes a description of both databases.

### 3.2.2. Automatic signature classification

In recent years, the state of the art in online signature verification has consistently relied on deep learning approaches. To demonstrate the competitiveness of our features, we integrated them into one of the best ASV, to the best of our knowledge. Specifically, we used the classifier proposed in [20], which employs deep representation learning through a convolutional recurrent adaptive network (CRAN). This sys- tem introduced the differentiable soft-DTW, incorporated it into the loss function, and developed an end-to-end trainable Deep soft-DTW (DsDTW) model that effectively combines CRAN with the traditional DTW mechanism.
To demonstrate the generalization of our features, it should be noted that we have not adjusted this ASV, but we have substituted the feature matrix for our $F_1$, $F_2$ and, $F_3$ with 2D and 3D robots.

### 3.2.3. Experimental protocol and metrics used

To facilitate fair comparisons with other studies, we adhered to established external protocols. The DeepSignDB database provides a standardized protocol based on skilled and random forgery experiments, with two scenarios: four signatures as references or one signature as a reference. We followed this protocol exactly as proposed. Additionally, the SVC-onGoing competition allows researchers to evaluate their systems at any time under consistent conditions. In this article, we utilized both resources to ensure a fair comparison of our proposed features and state-of-the-art methods by strictly adhering to the same experimental protocols. As in previous work with these databases, the EER was obtained in each case.

## 4. Analysis of robotic features in traditional machine-learning classifiers

The objective of this study was to analyse the performance of dynamic features – specifically, torque-based features – in ASV by using machine-learning classifiers. To achieve this, we examined the performance of kinematic features ($F_1$), dynamic features ($F_2$), and a combination of both ($F_3$) across two different ASV systems using multiple signature databases.

### 4.1. Analysis results of skilled forgeries

The first six columns of Table 5, present the equal error rates (EERs) in percentages for the skilled forgery experiment using DTW- and MAN-based ASVs with both 2D and 3D robots across various databases.

Table 5: Analysis of Skilled and Random Forgeries: evaluation of kinematic, dynamic, and combined features. Results in terms of EER (%).

| ASV | Database | Skilled forgeries | | | | | | Random forgeries | | | | | |
| | | 2D robot | | | 3D robot | | | 2D robot | | | 3D robot | | |
| | | F1 | F2 | F3 | F1 | F2 | F3 | F1 | F2 | F3 | F1 | F2 | F3 |
|-----|----------|------|------|------|------|------|------|------|------|------|------|------|------|
| DTW | MCYT-100 | 3.60 | 5.00 | 3.24 | 3.56 | 4.76 | 3.44 | 0.89 | 4.98 | 0.95 | 0.95 | 4.80 | 0.90 |
| | MCYT-330 | 3.79 | 7.70 | 4.73 | 3.90 | 7.53 | 4.96 | 0.89 | 15.05 | 2.61 | 0.83 | 15.35 | 2.80 |
| | BiosecurID | 2.71 | 3.22 | 2.21 | 2.59 | 3.28 | 2.40 | 1.17 | 5.30 | 1.17 | 1.24 | 5.99 | 1.38 |
| | Visual | 5.00 | 3.40 | 3.09 | 5.21 | 2.98 | 2.87 | 1.13 | 14.57 | 2.10 | 0.78 | 16.23 | 1.73 |
| | Blind | 2.84 | 4.55 | 3.41 | 2.61 | 4.55 | 3.30 | 0.26 | 10.38 | 1.31 | 0.09 | 10.51 | 0.78 |
| | SG-NOTE | - | - | - | - | - | - | 2.17 | 20.50 | 12.00 | 1.67 | 16.00 | 9.50 |
| | Bengali | - | - | - | - | - | - | 0.90 | 7.08 | 0.79 | 0.56 | 6.38 | 0.68 |
| | Devangari | - | - | - | - | - | - | 1.12 | 10.27 | 2.70 | 1.35 | 8.76 | 1.86 |
| | All together | 3.91 | 6.56 | 4.34 | 3.77 | 6.63 | 4.30 | 0.93 | 13.32 | 2.54 | 1.04 | 13.30 | 2.37 |
| MAN | MCYT-100 | 9.28 | 13.88 | 9.76 | 8.08 | 13.72 | 8.76 | 6.55 | 12.02 | 7.34 | 4.70 | 10.70 | 5.65 |
| | MCYT-330 | 9.33 | 14.55 | 10.41 | 8.47 | 13.77 | 9.30 | 6.61 | 12.96 | 7.79 | 5.08 | 11.83 | 6.40 |
| | BiosecurID | 7.13 | 7.32 | 6.50 | 5.05 | 7.26 | 5.73 | 8.40 | 12.12 | 8.67 | 6.05 | 12.32 | 6.95 |
| | Visual | 7.77 | 5.00 | 5.11 | 7.66 | 3.51 | 5.00 | 7.77 | 5.00 | 5.11 | 8.79 | 16.06 | 7.25 |
| | Blind | 7.05 | 8.64 | 4.66 | 8.07 | 6.25 | 5.57 | 4.30 | 12.72 | 4.47 | 4.85 | 12.10 | 4.56 |
| | SG-NOTE | - | - | - | - | - | - | 11.33 | 31.17 | 21.50 | 9.33 | 24.17 | 13.00 |
| | Bengali | - | - | - | - | - | - | 9.73 | 18.04 | 12.36 | 9.48 | 17.33 | 11.91 |
| | Devangari | - | - | - | - | - | - | 10.77 | 19.46 | 13.60 | 10.41 | 17.80 | 13.03 |
| | All together | 8.30 | 12.44 | 8.63 | 9.57 | 12.94 | 9.55 | 5.90 | 13.01 | 7.25 | 7.22 | 14.15 | 8.62 |

For the DTW-based ASV, similar performances were observed with the two robots across the databases. Specifically, the combination of features, $F_3$, yielded slightly better results with the 2D robot for some databases (e.g. MCYT-100, BiosecurID) and slightly better results with the 3D robot for others (e.g. Visual, Blind). The performance of $F_2$ was generally superior to that of $F_1$ in almost all cases, except for the Visual database. When combining features, i.e. $F_3$, it was found that, compared to $F_1$, the performances for MCYT-100, BiosecurID, and Visual improved with both robots. The best relative EER reduction was 38.2% (($5.00 - 3.09)/5.00 \times 100\%$) for Visual with the 2D robot and 44.9% (($5.21 - 2.98)/5.21 \times 100\%$) for Visual with the 3D robot. Conversely, the relative performance of $F_1$ deteriorated when combining features, particularly for MCYT-330, with a 24.8% decrease (($3.79 - 4.73)/3.79 \times 100\%$) for the 2D robot and a 27.2% decrease (($3.90 - 4.96)/3.90 \times 100\%$) for the 3D robot. To obtain an overall view of all databases, the scores were concatenated for performance analysis. In these cases, the best results were obtained with $F_1$, followed by $F_3$, with both robots. For the MAN-based ASV, performances were consistent across the databases with both robots. Overall, the performance with torques, $F_2$, was better than with $F_1$ for the Visual and Blind databases. The combination of kinematic and dynamic features did not significantly degrade performance. The worst combination damage was quantified as an 11.6% decrease (($9.33 - 10.41)/9.33 \times 100\%$) for MCYT-330 with the 2D robot and a 13.4% decrease (($5.05 - 5.73)/5.05 \times 100\%$) for BiosecurID with the 3D robot. However, significant improvements were observed when combining features for the SUSIG database, with a relative EER improvement of 34.2% (($7.77 - 5.11)/7.77 \times 100\%$) for Visual with the 2D robot and 34.7% (($7.66 - 5.00)/7.66 \times 100\%$) for the same database with the 3D robot. Again, a global perspective was obtained by concatenating the scores of all the databases to compute the EER. The combination of features, $F_3$, did not provide a clear conclusion as the performance improvements and degradations were very similar for the 2D and the 3D robot.

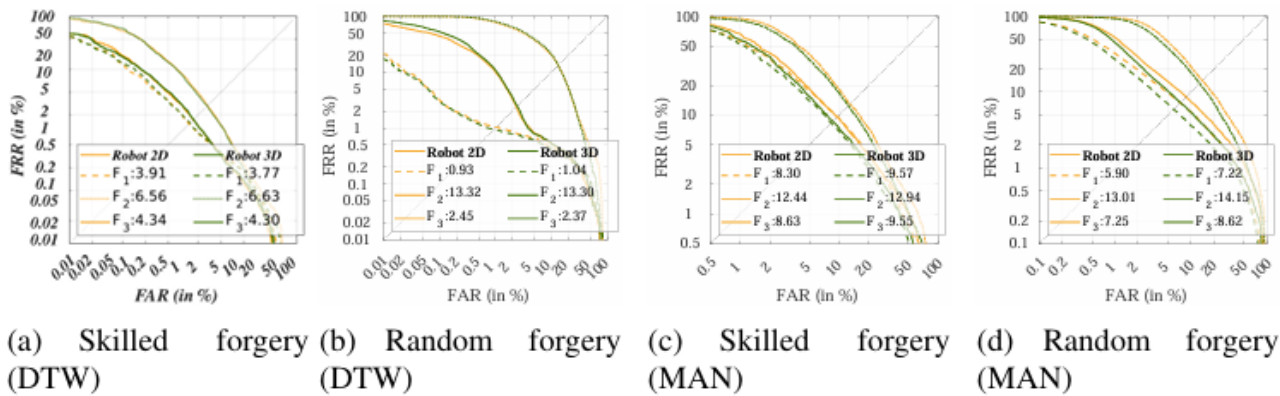

(a) Skilled forgery (DTW)  (b) Random forgery (DTW)  (c) Skilled forgery (MAN)  (d) Random forgery (MAN)

*Figure3: DET plots. Signature verification results in terms of EER with the DTW-based and Manhattan based ASVs when concatenating all scores.*

Finally, Figure 3-a and -c present the DET curves for the skilled forgery experiment with DTW- and MAN-based ASVs when all the scores were concatenated. In addition to the numerical analysis, it can be visually observed that the $F_1$ and $F_3$ curves are similar for both the robots across all the FAR and FRR values.

### 4.2. Analysis of the results of random forgeries

For the DTW-based ASV, the generalized coordinates demonstrated superior performance, as shown in the last six columns in Table 5. Notably, an EER of 0.09% in the Blind database was achieved with the 3D robot. The combination of features consistently resulted in positive effects compared to the use of torques alone. Thus, compared to $F_1$ and $F_3$, good results were observed with the 2D robot for the Bengali database and with the 3D robot for the MCYT-100 database.

For the MAN-based ASV, the combination of features, $F_3$, yielded better results for the Visual database with the 2D robot, and for both the Visual and Blind databases with the 3D robot. In other cases, the generalized coordinates achieved the best results, followed by the combination of features.

For a visual analysis, the DET curves in Figure 3-b and -d show that the curves are quite similar for both robots. However, for high FAR values, the features show similarities, while for lower FAR values dissimilarities are observed, especially for the DTW-based ASV. In this scenario, the generalized coordinates appear to be more competitive features for this classifier.

### 4.3. Calculation cost, execution time, and computational complexity

To evaluate the computational cost, the time complexity was assessed. For both robots, the generalized coordinates are determined by evaluating equations. Let $n$ be the length of a signature; the complexity to compute these coordinates is $O(n)$. For the 2D robot, computing the inverse dynamics also involves evaluating equations, resulting in a calculation cost of $O(n)$. However, for the 3D robot the algorithm used [37] formulates the inverse dynamics for a 3D robot with a generic number of joints. Let $n_j$ denote the number of joints in the 3D robot; the complexity to compute the torques is $cte = O(nj^6)$. Thus, the calculation cost to compute the torques for a signature is $O(cte \cdot n)$. Note that many previous studies used many of the features included in Table 1, and each of them have a calculation cost of $O(n)$.

Regarding the execution time, we conducted tests on the MCYT-330. Our code, implemented to calculate robotic features, was executed in Matlab version R2022b on an Intel(R) Core(TM) i9-10900F CPU at 2.80 GHz with 20 cores, running the Ubuntu 22.04.4 LTS operating system. For the 2D robot, the runtime for computing generalized coordinates were $0.04 \pm 0.02$ msec, and for computing torques, it was $0.03 \pm 0.01$ msec. For the 3D robot the runtime was $0.05 \pm 0.03$ msec for the generalized coordinates and $148.03 \pm 0.06$ msec for the torque-based features. Based on this analysis, we conclude that these execution times validate the applicability of our features in real use, which could potentially be optimized further in other programming languages, such as C, C++, or Python. Additionally, ASV systems on mobile devices could benefit from robotic features, as the runtime required to extract them demonstrates that the approach is efficient and practical for such applications. This is particularly relevant when deployed on cloud computing platforms, where computational resources can be dynamically scaled to meet demand.

Table 6: State-of-the-art performance in online automatic signature verification using the DeepSignDB dataset.

| Writing Input | Method | Skilled forg. | | Random forg. | | Average |
|---|---|---|---|---|---|---|
| | | 4vs1 | 1vs1 | 4vs1 | 1vs1 | |
| Stylus | DTW [20] | 4.53 | 7.06 | 1.23 | 1.98 | 3.70 |
| | TA-RNN [33] | 3.30 | 4.20 | **0.60** | **1.50** | 2.40 |
| | DsDTW (original) [20] | **2.54** | **4.04** | 0.97 | 1.69 | **2.31** |
| | $F_1$: 2D Robot + DsDTW | 3.80 | 5.87 | 2.06 | 2.82 | 3.64 |
| | $F_1$: 3D Robot + DsDTW | 3.59 | 5.21 | 1.87 | 2.22 | 3.22 |
| | $F_2$: 2D Robot + DsDTW | 3.69 | 5.82 | 1.98 | 2.82 | 3.58 |
| | $F_2$: 3D Robot + DsDTW | 3.46 | 5.21 | 1.78 | 2.52 | 3.24 |
| | $F_3$: 2D Robot + DsDTW | 3.71 | 5.74 | 1.89 | 2.48 | 3.46 |
| | $F_3$: 3D Robot + DsDTW | 3.58 | 5.40 | 1.71 | 2.37 | 3.27 |
| Finger | DTW [20] | 10.66 | 14.74 | 1.02 | **1.25** | 6.92 |
| | TA-RNN [33] | 11.30 | 13.80 | **1.00** | 1.80 | 7.00 |
| | DsDTW (original) [20] | 6.99 | 11.84 | 1.81 | 2.89 | 5.88 |
| | $F_1$: 2D Robot + DsDTW | 6.67 | 11.27 | 2.24 | 4.04 | 6.06 |
| | $F_1$: 3D Robot + DsDTW | 6.33 | 10.91 | 2.30 | 4.14 | 5.92 |
| | $F_2$: 2D Robot + DsDTW | 7.02 | 13.15 | 3.37 | 5.57 | 7.28 |
| | $F_2$: 3D Robot + DsDTW | 7.17 | 11.96 | 3.39 | 6.25 | 7.19 |
| | $F_3$: 2D Robot + DsDTW | **5.94** | **10.64** | 3.58 | 5.16 | 6.33 |
| | $F_3$: 3D Robot + DsDTW | 6.23 | 10.93 | 2.55 | 3.74 | **5.86** |

## 5. State-of-the-art results with deep-learning classification

For a fair comparison with state-of-the-art signature verification methods, a com- mon benchmark with a clear and standardized experimental protocol is essential [6].

Table 6 presents our results alongside those of previous works using the Deep- SignDB database. The best result in each category is highlighted in bold. For the finger modality, our features, when integrated into the deep-learning-based ASV [20], achieved the top position in three cases. We highlighted the best average value for $F_3$ with the 3D robot. Nevertheless, the results obtained indicate comparability with prior performance. These results demonstrate that our features can surpass the state of the art in several scenarios and achieve competitive performance when used within an external ASV.

Table 7: State-of-the-art performance in online automatic signature verification using the SVC2021 EvalDB dataset.

| Team | Method | Equal Error Rates (%) | | |
|---|---|---|---|---|
| | | Stylus | Finger | Styl/Fin |
| - | Baseline DTW | 13.08 | 14.92 | 14.67 |
| SIG | Online and offline approaches [16] | 7.50 | 10.14 | 9.96 |
| TUSUR-KIBEVS | Global features [14], CatBoost [30] | 6.44 | 13.39 | 11.42 |
| SigStat | Multiple distance scores | 11.75 | 13.29 | 14.28 |
| Mad-Lab | 1D version of ResNet-18 [17] | 9.83 | 17.23 | 14.21 |
| BiDA-Lab | TA-RNN [33] | 4.08 | 8.67 | 7.63 |
| DLVC-Lab | DsDTW (original) [20] | 3.33 | 7.41 | 6.04 |
| *This work* | $F_3$: 2D Robot + DsDTW | 3.75* | 8.56* | 8.88* |
| *This work* | $F_3$: 3D Robot + DsDTW | 4.00* | 8.88 | 10.02 |

*Top-3 state-of-the-art results

Finally, we participated in the SVC-onGoing competition using the combination of features, specifically $F_3$, extracted with the 2D and 3D robots, within the same deep- learning ASV [20]. The results are given in Table 7 with the SVC2021 EvalDB dataset and previous works. While the original ASV [20] achieved the best overall performance, our approach attained top-3 state-of-the-art results in many modalities. Overall, the results presented in this section demonstrate that our features are competitive for signature verification experiments.

## 6. Conclusion

### 6.1. Key findings

This paper proves the hypothesis that robotic features are useful for ASV systems. We applied a Lagrangian formulation to derive novel information for the purpose of verifying signatures, which consisted of generalized coordinates and torques. To ex- tract these features, a 2D robot and a 3D robot were proposed. Through a series of extensive experiments using multiple databases and two distinct automatic signature verifiers, we showcase an analysis of the robotic features in online signature verification with traditional machine-learning classifiers. Furthermore, our results showed that integrating these features into a deep-learning ASV system can surpass current state-of-the-art performance.

One advantage of using robotic features, compared to more black-box approaches, is that the integration of kinematic and dynamic features introduces additional physical information into the verification process. These features establish a relationship between torques (dynamic features) and generalized coordinates and velocities (kinematic features), which are derived from the parametric equations of the signatures. This makes the features more interpretable, and particularly valuable for forensic applications. Furthermore, while the kinematics of movement has been explored in this field, this work advances the use of dynamics. This represents a significant step forward in the intersection of pattern recognition and applied physics, and it could provide a basis for further research in this direction.

### 6.2. Limitations

One advantage of the Lagrangian formulation is its flexibility in modifying the physical parameters that represent the signer, such as masses, lengths, and inertial values. In this study, we kept these values fixed for all signers. We conducted a preliminary test by changing the physical values for each signer, but observed no significant improvements. A potential direction for individualized parameterization could involve using machine-learning techniques to establish relationships between the body dimensions (such as arm length, mass, and inertia) and the signature trajectory and pressure. Thus personalizing the physical values of the writers remains a limitation in this work and presents a promising area for future research.

### 6.3. Future works

This study highlights the potential of the Lagrangian formulation for characterizing online signatures, paving the way for further advancements in the field of ASV systems. However, there are other possibilities to explore, such as combining robotic features with additional features, as shown in Table 1, which could further enhance performance.

Beyond signature recognition, it is worth investigating the use of estimated dynamic features from robots in other behavioral biometric traits, such as mouse movements, touchscreen behavior, or keystroke patterns. For example, gait recognition could benefit from these principles, as the Lagrangian formulation has already demonstrated its effectiveness in modelling walking movement [27].

Table A.8: Influence of writing on a horizontal or vertical plane (performance in EER (%)).

| Database | ASV | Skilled Forg. $F_2$ | $F_3$ | Random Forg. $F_2$ | $F_3$ | |
|---|---|---|---|---|---|---|
| MCYT-100 | DTW | 5.08 | 3.28 | 5.09 | 0.95 | H |
| | | 5.00 | 3.24 | 4.98 | 0.95 | V |
| | MAN | 14.16 | 9.72 | 12.14 | 7.34 | H |
| | | 13.88 | 9.76 | 12.02 | 7.34 | V |
| MCYT-330 | DTW | 7.84 | 4.80 | 15.64 | 2.86 | H |
| | | 7.70 | 4.73 | 15.05 | 2.61 | V |
| | MAN | 14.72 | 10.42 | 12.92 | 7.77 | H |
| | | 15.55 | 10.41 | 12.96 | 7.79 | V |
| Biose-curID | DTW | 3.28 | 2.34 | 5.51 | 1.10 | H |
| | | 3.22 | 2.21 | 5.30 | 1.07 | V |
| | MAN | 7.45 | 6.69 | 12.12 | 8.61 | H |
| | | 7.32 | 6.50 | 12.12 | 8.67 | V |
| Visual | DTW | 3.40 | 3.30 | 14.99 | 2.26 | H |
| | | 3.40 | 3.09 | 14.57 | 2.10 | V |
| | MAN | 4.89 | 5.11 | 17.94 | 8.36 | H |
| | | 5.00 | 5.11 | 17.80 | 8.27 | V |
| Blind | DTW | 4.55 | 3.64 | 10.79 | 1.52 | H |
| | | 4.55 | 3.41 | 10.38 | 1.31 | V |
| | MAN | 8.41 | 4.66 | 13.02 | 4.47 | H |
| | | 8.64 | 4.66 | 12.72 | 4.47 | V |
| Benga-li75 | DTW | | | 7.86 | 0.92 | H |
| | | | | 7.08 | 0.79 | V |
| | MAN | | | 18.25 | 12.36 | H |
| | | | | 18.04 | 12.36 | V |
| Devana-gari75 | DTW | | | 11.50 | 2.94 | H |
| | | | | 10.27 | 2.70 | V |
| | MAN | | | 19.37 | 13.64 | H |
| | | | | 19.46 | 13.60 | V |
| mobile SG-NOTE | DTW | | | 20.50 | 12.00 | H |
| | | | | 20.50 | 12.00 | V |
| | MAN | | | 31.17 | 21.50 | H |
| | | | | 31.17 | 21.50 | V |

*H:* Horizontal, without potential energy

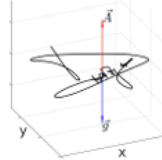

*V:* Vertical, with potential energy

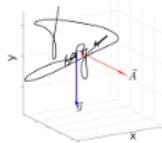

## 7. Acknowledgments

This research was partly supported by the PID2023-146620OB-I00 project, funded by MICIU/AEI 10.13039/501100011033 and the European Union's ERDF program, and partly by the CajaCanaria and la Caixa (2023DIG05).

### 7.1. Appendix A. Impact of horizontal and vertical writing with the 2D robot

We explored two robot pose options for the 2D robot. Firstly, we ignored the potential energy. In this configuration, the vector area ($\vec{A}$) of the horizontal plane and gravity ($\vec{g}$) were parallel. Secondly, we considered the potential energy when the robot wrote on a vertical plane. Here, the area and gravity vectors were perpendicular. The numerical results for both cases are shown in Table A.8.

For this, we evaluated DTW-based and MAN-based ASVs using $F_2$ and $F_3$ as fea- tures. Numerically, although the MAN-based ASV sometimes exhibited slightly better performance when writing on a vertical surface, the DTW-based ASV consistently per- formed slightly better for both skilled and random forgeries.
In reality, arm's positioning during signing is neither wholly horizontal nor ver- tical. In this real scenario the potential energy would contribute to the Lagrangian formulation. To account for this source of energy, and due to the slightly better results obtained, we assumed that the writing was done on a vertical plane for the remainder of our experiments.